\documentclass[11pt,a4paper]{article}
\usepackage[table]{xcolor}
\usepackage{multirow}
\usepackage[hyperref]{acl2019}
\usepackage{times}
\usepackage{latexsym}
\usepackage{url}
\usepackage{graphicx}
\usepackage{booktabs}

\aclfinalcopy 

\newcommand{\sectab}{\cellcolor{gray!40}}

\title{Unsung Challenges of Building and Deploying Language Technologies for Low Resource Language Communities}

\author{ \textbf{Pratik Joshi\textsuperscript{1}} \quad \textbf{Christain Barnes\textsuperscript{2}\thanks{Work done during internship at Microsoft Research}} \quad \textbf{Sebastin Santy\textsuperscript{1}} \quad \textbf{Simran Khanuja\textsuperscript{1}}  \\ \textbf{Sanket Shah\textsuperscript{1}} \quad \textbf{Anirudh Srinivasan\textsuperscript{1}} \quad \textbf{Satwik Bhattamishra\textsuperscript{1}} \\ \textbf{Sunayana Sitaram\textsuperscript{1}} \quad \textbf{Monojit Choudhury\textsuperscript{1}} \quad \textbf{Kalika Bali\textsuperscript{1}\thanks{Email: kalikab@microsoft.com}} \\
\textsuperscript{1} Microsoft Research, Bangalore, India \\
\textsuperscript{2} Stanford University \\
}

\date{}

\begin{document}
\maketitle
\begin{abstract}
  In this paper, we examine and analyze the challenges associated with developing and introducing language technologies to low-resource language communities. While doing so, we bring to light the successes and failures of past work in this area, challenges being faced in doing so, and what they have achieved. Throughout this paper, we take a problem-facing approach and describe essential factors which the success of such technologies hinges upon. We present the various aspects in a manner which clarify and lay out the different tasks involved, which can aid organizations looking to make an impact in this area. We take the example of Gondi, an extremely-low resource Indian language, to reinforce and complement our discussion. 
\end{abstract}

\newcommand{\ques}[1]{\textbf{#1}\\}

\section{Introduction}



Technology pervades all aspects of society and continues to change the way people access and share information, learn and educate, as well as provide and access services. Language is the main medium through which such transformational technology can be integrated into the socioeconomic processes of a community. Natural Language Processing (NLP) and Speech systems, therefore, break down barriers and enable users and whole communities with easy access to information and services. However, the current trend in building language technology is designed to work on languages with very high resources in terms of data and infrastructure.

\begin{figure}
	\centering
	\includegraphics[scale=0.4]{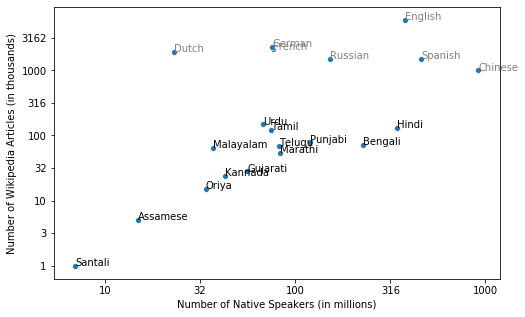}
	\caption{The points represent the disparity between number of wikipedia articles in comparison with the number of native speakers for a particular language.}
	\label{fig: langvsspeakers}
\end{figure}

Also, as Machine Learning (ML) and NLP practitioners, we get caught up in an information-theoretic view of the problem, e.g., focusing on incremental improvements of performance on benchmarks or capturing accurate distributions over data, and tend to forget that the raison d'être of NLP is to build systems that add value to its users \cite{ruder_2019}. We want to build models that enable people to read the news that was not written in their language, ask questions about their health when they do not have access to a doctor, etc.  And while these technology applications are more and more ubiquitous for languages with a lot of data, a larger majority of languages remain resource-poor and bereft of such systems. As discussed in the United Nations e-government survey \cite{nations2014united}, “one of the most important obstacles to e-inclusion, particularly among vulnerable groups with little education, is language”. Thus, by excluding these languages from reaping the benefits of the advancements in language technology, we marginalize the already vulnerable groups even further.

India is a highly multilingual society and home to some of the largest language communities in the world. 6 out of 20 most-spoken (native) languages in the world are Indic. Ethnologue \cite{simons2017ethnologue} records 461 tongues in India out of 6912 worldwide  (6\%), the 4th largest belonging to any single country in the world. 122 of these languages are spoken by more than 10,000 people. 29 languages have more than 1 million speakers, which include indigenous tribal languages like Gondi and Mundari, some without a supported writing system or script. Despite the large numbers of users, most of these languages have very little data available. Figure \ref{fig: langvsspeakers} shows that as compared to some of the much lesser spoken languages like German, Indic languages are severely low resourced. In a vast country like India, access to information thus becomes a huge concern. This lack of information means that not only do these communities not have information in domains like agriculture, health, weather etc., which could improve their quality of lives, but they may also not be aware of their basic rights as citizens of the country.

In this paper, we take the position that the current direction of advanced language technology towards extremely high data requirements can have severe socio-economic implications for a majority of language communities in the world. We focus on specific aspects of designing and building systems and applications for low resource languages and their speech communities to exemplify viable social impact through language technology. We begin by discussing the aspect of information exchange, which is the core motivation behind enabling low-resource language communities. We then steer our analysis towards the design and creation of an interface for people in these communities to simplify and enrich the process of information exchange. Finally, we gather insights about how to deploy these technologies to ensure extensive impact by studying and taking inspiration from existing technological deployments.

We use Gondi, a South-Central Dravidian language in the vulnerable category on UNESCO's Atlas of the Worlds Languages in Danger \cite{moseley2010atlas}, as an example wherever possible. Spoken by nearly 3 million people \cite{census2011} in the Indian states of Chhattisgarh, Andhra, Odisha, Maharashtra and Karnataka, it is heavily influenced by the dominant state language. However, it is also one of the least resourced languages in India, with very little available data and technology.

We believe that the components discussed in the sections below encapsulate the spectrum of issues surrounding this field and that all future discussions in this area will also fall under the umbrella of these categories. We believe that by focusing on Gondi, we will not only empower the Gondi community but more importantly, understand and create a pipeline or framework which can serve as a clear guide for potential ventures which plan on introducing disruptive language technologies in under-served communities.

\section{Information Exchange}
The primary element in communication is information exchange. People living in less connected areas are often unable to get the kind of information they need, due to various socio-economical and technological barriers. As a result, they miss out on crucial knowledge required to improve their well-being. There are three co-dependent aspects woven into the fabric of information exchange - access of information, quality and coverage of the information and methods to create and digitize available knowledge (generation).



\subsection{Access}
 This section refers to past work and current ventures of making digital resources adequately available and accessible to people. 
 
\subsubsection{Making information accessible to people}

Less-connected and technologically underdeveloped areas often suffer from the limited accessibility of up-to-date information. Providing more individuals access to the online repositories of information can often help them improve their well-being.  

There are some situations particularly during natural calamities where the absence of notifications about potentially disaster-prone areas can result in life and death situations of individuals. People in regions with sparse connectivity often fall victim to these incidents due to lack of timely updates. Using technical platforms to support the spread of information to these regions is an important goal to keep in mind. LORELEI \cite{strassel2016lorelei} is a DARPA funded initiative with the goal of the building of technologies for dealing and responding to disasters in low resource language communities. Similar initiatives in India would be capable of saving lives.

The daily function and health of individuals in a community can be influenced positively by the dissemination of relevant information. For example, healthcare and agricultural knowledge can affect the prosperity of a rural household, making them aware of potential solutions and remedies which can be acquired. There has been a considerable body of work focused on technology for healthcare access, which includes telemedicine \cite{telemedicine} and remote diagnosis \cite{diagnosis}. While the use of telecenters to spread information on agricultural practises has been employed, persuading users to regularly use the telecenters \cite{telecenter} is a challenge, which could be addressed by the use of language technologies to simplify access. VideoKheti \cite{cuendet2013videokheti} is an example of a voice-based application which provides educational videos to farmers about effective agricultural practices. Similar studies have been carried out to assess the effectiveness of voice-activated applications for farming \cite{patel2010avaaj}. There are considerable challenges, however, to ensure that these solutions are inclusive and accessible to low-literate and less-connected users.

Similarly, there are situations where there are certain rights and duties which an individual as a citizen of India is entitled to. Some communities have long been exploited and ill-treated \cite{ganguly2003forest}, and providing them information regarding their rights as well as accurate news could foster a sense of solidarity within the community and encourage them to make their voice heard. An extensive study on the impact of CGNet Swara \cite{marathe2015revisiting} showed that this citizen journalism platform inspired people in rural communities, gave them a feeling of being heard, and provided a venue to voice their grievances. There are also other promising ventures such as Awaaz De \cite{patel2010avaaj} and Gram Vaani \cite{moitra2016design} which aim to boost social activism in a similar manner.




\subsubsection{Making more digital content available}
The process of enabling more low-resource language communities with tools to access online information alone is not sufficient. There need to be steps taken to make more of the content which exists online interpretable to people in these communities. For example, The Indian Constitution and other similar official communications from the government are written in 22 scheduled languages of India. Lack of access to other related documents deprives them of basic information. This is where building robust machine translation tools for low resource languages can help. Cross-language information retrieval makes extensive use of these translation mechanisms \cite{zhou2012translation} where information is retrieved in a language different from the language of the user's query. \citet{mcnamee2002comparing} describes a system making use of minimal resources to perform the same.

There is huge potential for language technologies to be involved in content creation and information access. Further, more accurate retrieval methods can help the user get relevant information specific to their needs and context in their own language.

\subsubsection{Making NLP models more accessible to low resource languages}
Often, many state-of-the-art tools cannot be applied to low-resource languages due to the lack of data. Table \ref{tab: nlptable} describes the various technologies and their presence concerning languages with different levels of resource availability and the ease of data collection. We can observe that for low resource languages, there is considerable difficulty in adopting these tools. Machine Translation can potentially be used as a fix to bridge the gap. Translation engines can help in translating documents from minority languages to majority languages. This allows the pool of data to be used in a number of NLP tasks like sentiment analysis and summarization. Doing so allows us to leverage the existing body of work in NLP done on resource-rich languages and subsequently apply it to the resource-poor languages, thereby foregoing any attempt to reinvent the wheel for these languages. This ensures a quicker and wider impact.\citet{wan2008using} performs sentiment analysis on Chinese customer reviews by translating them to English. They observe that the quality of machine translation systems are sufficient for sentiment analysis to be performed on the automatically translated texts without a substantial trade-off in accuracy.

\begin{table*}
\begin{tabular}{|l|l|l|l|l|l|l|p{1.4cm}|}
\hline
\multicolumn{1}{|c|}{Technology} & \multicolumn{4}{p{5cm}|}{\centering Availability of technology for the resource status of a language}                          &  \multicolumn{3}{p{5.25cm}|}{\sectab Data/Expertise Requirement}                                                \\ \hline
\multicolumn{1}{|c|}{}           & \multicolumn{1}{p{1.2cm}|}{\centering High} & \multicolumn{1}{p{1.4cm}|}{\centering Moderate} & \multicolumn{1}{p{1.2cm}|}{\centering Low} & \multicolumn{1}{p{1.2cm}|}{\centering No} & \multicolumn{1}{p{1.5cm}|}{\sectab Linguistic Expertise} &  \multicolumn{1}{p{1.5cm}|}{\sectab Unlabeled Data}  & \sectab Labeled Data \\ \hline
\multicolumn{8}{|c|}{\cellcolor{gray} Input/Output Support}                                                                                                                                                                                                         \\ \hline
Font \& Keyboard                 & \multicolumn{1}{p{1.2cm}|}{$\star \star \star$}     & \multicolumn{1}{p{1.4cm}|}{$\star \star \star$}         & \multicolumn{1}{p{1.2cm}|}{$\star \star \star$}    & \multicolumn{1}{p{1.2cm}|}{$\star \star $}   & \sectab $\star \star \star$                     & \sectab           &     \sectab         \\ \hline
Speech-to-Text                   & $\star \star \star$ & $\star \star$  & \cellcolor{gray!75} & \cellcolor{gray!75}  &  \sectab $\star$ &  \sectab $\star \star$  & \sectab $\star \star \star$  \\ \hline
Text-to-Speech                   & $\star \star \star$ & $\star \star $ & $\star $ & \cellcolor{gray!75} & \sectab $\star \star \star$ & \sectab & \sectab $\star \star $ \\ \hline
Text Prediction                  & $\star \star \star$ & $\star \star \star$ & $\star \star$ & \cellcolor{gray!75} & \sectab & \sectab $\star \star \star$ & \sectab \\ \hline
Spell Checker                    & $\star \star \star$ & $\star \star \star$ & $\star \star$ & \cellcolor{gray!75} & \sectab $\star \star \star$ & \sectab $\star \star $ & \sectab \\ \hline
Grammar Checker                  & $\star \star \star$ & $\star \star $ & \cellcolor{gray!75} & \cellcolor{gray!75} & \sectab $\star \star $ & \sectab $\star \star \star$ & \sectab $\star \star $ \\ \hline

\multicolumn{8}{|c|}{\cellcolor{gray} Local Language UI}                                                                                                                                                                                                            \\ \hline
                                 & $\star \star \star$ & $\star \star \star$ & $\star \star $ & \cellcolor{gray!75} & \sectab $\star \star \star$ & \sectab & \sectab \\ \hline

\multicolumn{8}{|c|}{\cellcolor{gray}Information Access}                                                                                                                                                                                                           \\ \hline
Text Search                      & $\star \star \star$ & $\star \star $ & $\star$ & \cellcolor{gray!75} & \sectab $\star$ & \sectab $\star \star \star$ & \sectab $\star \star$ \\ \hline
Machine Translation              & $\star \star $ & $\star$ & $\star$ & \cellcolor{gray!75} & \sectab $\star \star $ & \sectab  & \sectab $\star \star \star$ \\ \hline
Voice to Text Search             & $\star \star \star$ & $\star $ & \cellcolor{gray!75} & \cellcolor{gray!75} & \sectab & \sectab $\star$ & \sectab $\star \star \star$ \\ \hline
Voice to Speech Search           & $\star \star$ & $\star$ & \cellcolor{gray!75} & \cellcolor{gray!75} & \sectab $\star $ & \sectab $\star \star \star$ & \sectab $\star \star \star$ \\ \hline

\multicolumn{8}{|c|}{\cellcolor{gray} Conversational Systems}                                                                                                                                                                                                       \\ \hline
                                & $\star \star $ & $\star $ & \cellcolor{gray!75} & \cellcolor{gray!75} & \sectab $\star \star \star$ & \sectab $\star \star \star$ & \sectab $ \star \star \star$ \\ \hline
\end{tabular}
\caption{Enabling language technologies, their availability and quality ( $ \star \star \star $ - excellent quality technology, $ \star \star $ - moderately good but usable, $ \star $ - rudimentary and not practically useful) for differently resourced languages, and their data/knowledge requirements ($ \star \star \star $ - very high data/expertise, $ \star \star $ - moderate, $ \star $ - nominal and easily procurable). This information is based on authors' analysis and personal experience.}
\label{tab: nlptable}
\end{table*}



\subsection{Generation}
This section refers to the generation of digital content which enriches online repositories with more diverse sets of information.

\subsubsection{Digitization of Documents}
There is a need to generate digital information and content for low-resource languages. It not only benefits the community by creating digital content for their needs, but it also provides data which can be used to train data-driven language technologies, such as ASRs, translation systems, and optical character recognition systems. Efforts to digitize content in India have been conducted in the past few years. The Government of India launched the Digital India initiative \footnote{http://www.digitalindia.gov.in/} in 2015, which aims to digitize government documents in one of India's 120+ local languages. Such initiatives have evidently been useful before. For instance, the IMPACT project \footnote{http://www.impact-project.eu/} by the European Union was a large scale digitization project which helped push a lot of innovative work towards OCR and language technology for historical text retrieval and processing. IMPRINT is a similar initiative created by the Ministry of Human Resource Development (MHRD) to drive further research towards addressing such challenges.

The recent advancements in OCR technologies can propel efforts to digitize more handwritten documents. Such initiatives are already being undertaken to digitize and revive historical languages in Japan \cite{tkasasagi}. Digital India library is a project that aims towards digitizing books and making them available online. Apart from printed books, a lot of ancient literature is written on palm leaves. The Regional Mega Scan Centre (RMSC) at IIIT Hyderabad has digitized over 100,000 books, one-third of which are in Indian Languages and additionally, they have also digitized text from scans of palm leaves. More initiatives such as these will help preserve and revive a number of languages that are part of the Indian heritage.

\subsubsection{Crowdsourcing}
Data collection via crowdsourcing can be a challenge for low resource languages, primarily due to the expensive nature of the task coupled with the lack of commercial demand for such data. Thus, collecting this data at low cost becomes an important priority. Project Karya is a crowdsourcing platform which provides digital work to low-income workers. Although the data quality can be a concern, promising results have shown otherwise. \citet{chopra2019exploring} tested the quality of crowdsourced data in rural regions of India, tasking individuals with the digitization of Hindi/Marathi handwritten documents. A 96.7\% accuracy of annotation was yielded, proving that there is potential in this area. Recently, collection of Marathi speech data is also being conducted. In a similar fashion, Navana Tech\footnote{https://navanatech.in/}, a startup, has been collecting data in mid and low-resource languages of verbal banking queries so that they can be integrated into various banking application platforms for financial inclusion. Such crowdsourcing platforms not only act as a potential data for low-resource communities, they also benefit low-income workers by increasing their current daily wage. Such ventures would enhance the inclusion of such workers in the digitization process, something which aligns with the aims of the Digital India mission. 

The collection of data in an extremely low-resource language like Gondi can be particularly tricky, additionally considering the fact that Gondi does not have an official script. Pratham Books \footnote{https://prathambooks.org/} is a non-profit organization which aims to democratize access to books for children. They recently hosted a workshop where they trained members of the local community to translate books on StoryWeaver\footnote{https://storyweaver.org.in/}, their open-source publication platform. At the end of this workshop, approximately 200 books were translated from Hindi to Gondi (Devanagiri script). This was the first time children's books were made available in Gondi, and it also sparked the creation of parallel data for Hindi-Gondi translation systems.

\section{Interface}
The design of a user-friendly interface plays a very crucial role in ensuring that the deployed technology encompasses all strata of society. It is often seen that a majority of target users have not had the privilege of education, and show varying levels of literacy, both foundational and digital. In such scenarios, text-based modalities pose several limitations from both the user and designer perspectives, and graphical user interfaces have been the preferred choice in these applications. \citet{thies2015user} reports that text-based interfaces were completely redundant for illiterate users and severely error-prone for literate but novice users. Further, several languages do not have unique keyboard standards or fonts, and some do not have a script at all \cite{boyera2007mobile}.

To overcome these issues with text, speech as a modality has also been deployed with varying success. `CGNet Swara', a citizen-run journalism portal, uses a phone-based IVR system to educate illiterate users \cite{mudliar2013emergent}. ’Avaaj Utalo’ allows users to make simple phone calls to ask questions or browse questions and answers asked on agricultural topics \cite{patel2010avaaj}. ’Spoken Web’ is another application wherein users can create ’voice sites’ analogous to ’websites’ which can then be easily accessed through voice interaction on mobile phones \cite{kumar2010spoken}. These serve to provide farmers with relevant crop and market information.
An attempt to leverage the complementarity of voice and graphic-based inputs was made by VideoKheti, a mobile system with a multi-modal interface for low-literate farmers providing agricultural extension videos on command in their own language or dialect \cite{cuendet2013videokheti}. They report that people in these communities find it difficult to use softkey type keyboards that are extremely common on modern smartphones. Instead, they proposed a system comprising of large buttons, graphics and some voice input. Such a system for delivering information to farmers was made and they showed that the farmers were very comfortable using it. Their results also show that a speech interface alone was not enough for that scenario, except in cases where the search list was long and the results were dependent on keywords or short phrases. Similarly, the Adivasi Radio App \footnote{\href{https://play.google.com/store/apps/details?id=com.anuragShukla06.AdivasiRadio}{AdivasiRadio - Google Play}}, based on text-to-speech (TTS) technology, is developed to read out written reports in Gondi, one of the main tribal languages in Chhattisgarh. Bolo is another mobile application which uses a very simple interface to improve children's literacy in India. 
 Project Karya also proposes to divide massive digital tasks into ”microwork” and crowdsource this work to millions of people in rural India via phones \footnote{\href{https://www.microsoft.com/en-us/research/video/project-karya-giving-dignified-digital-work-people-powered-bulk-data-networking/}{Project Karya}}. 

 While voice might solve the foundational literacy problems, the lack of digital literacy is often more challenging to overcome. \cite{mondal2019} demonstrate the use of an app to teach the Mundari language to children. The app comprised of a series of games designed with the help of the community. The content was delivered in the Bangla script, which was what the children were taught in school. Their study noted that children from such communities found the usage of a smartphone to be difficult.

Relying on voice-based systems also poses a few challenges. It is not easy to build robust ASR systems for these languages due to severe lack of data, dialect variations and several such constraints. An attempt to resolve this was made with the development of the SALAAM ASR \cite{qiao2010small} which uses the acoustic model of an existing ASR and performs a cross-lingual phoneme mapping between the source and target language. This, however, is limited to recognition of a very small set of vocabulary, but finds use due to its' cost-effective and low resource setting. 

\section{Deployment and Impact}
After developing technologies to provide information, and ensuring that the applications are designed in such a way that they are accessible to the population, the technology must be effectively deployed. Specialized applications are useless if they are not deployed properly in a way that accesses their end-users. When deploying a specially developed technology, the application must be deployed with consideration of the existing community dynamics. For any deployment to be successful, it usually must be able to be purposefully integrated into the lifestyle of community members - or have strong utilization incentives if it is a transformative technology. In this section, we will review examples of technology dissemination to low-resource/rural communities, and the impacts of effective deployment. While some technologies that we examine are not deployed utilizing low-resource languages specifically, the types of rural communities and villages in which they are deployed are analogous to the contexts in which low-resource languages exist, and clear parallels can be drawn.

Integrating the usage of a language technology intervention into a community in a low-resource context requires much more simply introducing the technology. Unlike hardware interventions and innovations like solar panels or new agricultural tools, language technologies often rely on the delivery, exchange, and utilization of information, which is much less tangible than physical solutions. This is especially for people with limited previous exposure to digital technology. Upon observing a selection of language-based interventions that were deployed in low-resource contexts, we observed that the most successful deployments of technologies tended to have three components of success. They:  1) Initially launched by seeding with target communities, 2) Worked closely to engage the community itself with the technology and information, and 3) Provided a strong incentive structure to adapt the technology - this incentive could be as simple as payments or as complex as communicated benefits from the technology.  

\subsection{Case Studies}
In this section, we will be reviewing and comparing three separate technological systems, Learn2Earn \cite{theis2019}, Mobile Vaani \cite{moitra2016design}, and the Climate and Agriculture Information Service (CAIS)\cite{Christensen:2019:YMC:3287098.3287109}, and see how they utilized the rules of successful deployment outlined above. Learn2Earn, developed by Microsoft Research, is a simple IVR based mobile language technology app which uses quizzes to educate people and spread public awareness campaigns, launched initially in rural central India. Mobile Vaani is a large-scale and broad community-based IVR media exchange platform, developed by the NGO Gram Vaani. It currently has over 100,000 unique monthly active users, and processes 10,000 calls per day across the three Indian states of Bihar, Jharkhand, and Madhya Pradesh. Finally, the CAIS system is an SMS-based information delivery system designed for farmers who live in a rural, low-resource and no connectivity agricultural village on the Char Islands in the Bangladeshi Chalan Beel Wetland. This application provides weather data and agricultural advice to farmers on a periodic basis and was developed by a collaboration between mPower and two local NGOs. 
After designing the platform in accessible ways, each deployment process began with the seeding within a target community within itself. All examples that we studied became successful only after a small scale launch of their product. These launches occurred in different ways but were all based on targeting a starting group of users and incentivizing them to utilize and share the product. The initial users were people who were somewhat fluent in the technology (either through training or existing knowledge), and who knew of or had specialized needs that the technology could address.
\subsubsection{Learn2Earn}
Learn2Earn was built as a tech-enabled information dissemination system; its original information awareness campaign centred on informing farmers about their rights as guaranteed in India’s Forest Rights act. Because of the nature of their message content and delivery, the researchers decided to seed the platform with a single advertisement on an existing IVR channel already utilized by farmers. This advertisement reached 150 people, and provided them with distinct financial incentives to both call the platform, and invite friends to the platform. While only 17 of the original listeners of the advertisement went on to call the number, those respondents were members of the relevant community (farmers who were familiar with IVR technology) and were networked through family and friendships to additional ideal users of the platform. Within 7 weeks, the incentive structure allowed the platform to spread from the original 17 users to over 17,000, with little influence from the platform respondents. \cite{theis2019} 
\subsubsection{Mobile Vaani}
Mobile Vaani initially tried to launch in 2011 by encouraging employees from their partner NGOs to distribute their platform. The platform was initially imagined as a voice-based, inclusive medium for communities to express their grievances and communicate with each other digitally. The initial employees who recruited for the platform were not from the community but did work closely with them regularly. While there was some initial success in the launch, the mobile Vaani Team were unable to grow at a significant pace because they informed the end-users about their intended design and usage of the technology, which “set unrealistic expectations of the platform in the minds of the participating users” after the technology could not be used in the exact way that it was encouraged.  A few months later, the platform decided to re-launch and expand by recruiting a series of trained and compensated volunteers from a variety of communities that they hoped to engage. During the second “launch,” the community members were able to learn about the platform, and adapt it to their specific use cases. The platform began to gain popularity during a teacher’s strike in the state of Jharkhand – where a specific use case for expressing grievances powered by the community arose. 
\subsubsection{CAIS}
The CAIS platform launched in direct collaboration with the NGO partner for the village – every available farmer registered their name, number, and crop type with the NGO partner – and consequently the target population was integrated from the start. As the programs grew, each engaged with the community on a high level.  In the case of all three platforms, a specific population, and a very specific understanding of that population’s needs had to be identified before the platform could be relatively effective.
Even after the deployment of the platform, care and close integration with community systems had to be done. The village in which CAIS worked had a system of self-empowerment groups, that had been organized by the NGO. Each group had a leader; and while not every villager in each group had a basic phone, every group leader did. Consequently, the researchers behind CAIS worked to ensure that every group leader was engaged in with CAIS and that they would relay their CAIS informational updates to the villagers that they lead. Similarly, the CAIS researchers worked closely with village leaders to determine who was able to access the information in SMS form and deployed informational physical posters as a substitute to those portions of the village population who could not. This intensive work led to the successful adaptation of technology to the benefit of the farmer’s yields. The researchers behind the Vaani system also continued to expand the system through a local network of volunteers. \cite{moitra2016design}
\subsubsection{CGNET Swara}
CGNet Swara, which we introduced earlier in this paper, also increased their initial participation by engaging with the wider community by holding in-person training and awareness sessions. They have conducted over 50 workshops, and have trained more than 2,000 members of various communities \cite{marathe2015revisiting}. Outreach activities such as these also allowed an increased spread of awareness via word-of-mouth. From these examples, it is clear to see that community engagement is the absolute key to spreading technology. 
Incentives –both monetary and situational – are a huge way that these platforms were able to engage their initial users. Incentives served to empower individuals to become champions of the platform and increased the enabled them to use their knowledge of the community and existing peer networks to deliver the technology where it was needed. All platforms used incentives of some sort; Learn2Earn used a direct payment for recruitment + participation, and also delivered relevant topics to the users. Mobile Vaani provided financial incentives to the volunteers who mobilized to evangelize the product. CAIS did not provide monetary incentives but instead brought technology that had an actionable and tangential impact on the daily lives of farmers. With the deployment of these technologies, direct needs of the population were solved.

\section{Conclusion}

The boost in recent advancements in NLP research has started breaking down communication and information barriers. This, coupled with in-depth studies on the socio-economic benefits of enabling less-connected communities with technology, provides a strong argument for increasing investment in this area. It is promising to observe increased innovation and steady progress in the empowerment of rural communities using language tools. Increased exposure to the challenges and works in this area can catalyse developments in improving inclusion and information dissemination. We hope that this paper will provide pointers in the right direction for potential ventures that plan on introducing disruptive language technologies to marginalized communities.

\bibliography{acl2019}
\bibliographystyle{acl_natbib}

\appendix

\end{document}